%% file: main.tex
\documentclass[runningheads]{llncs}
\usepackage[T1]{fontenc}
\usepackage{graphicx}
\usepackage{adjustbox}
\usepackage{caption}
\usepackage{subcaption}
\usepackage{makecell}
\usepackage{tikz}
\def\checkmark{\tikz\fill[scale=0.4](0,.35) -- (.25,0) -- (1,.7) -- (.25,.15) -- cycle;} 
\usepackage{diagbox}

\usepackage{amsmath}

\begin{document}
\title{Learning Representations for Masked Facial Recovery}
% \thanks{Supported by organization x.}}
%
\titlerunning{Learning Representations for Masked Facial Recovery}
\author{Zaigham A. Randhawa \and Shivang Patel \and
Donald A. Adjeroh \and \\
Gianfranco Doretto}
\authorrunning{Z. A. Randhawa et al.}
\institute{West Virginia University, Morgantown, WV 26506, USA\\
\email{\{zar00002,sap00008,daadjeroh,gidoretto\}@mix.wvu.edu}}
\maketitle              % typeset the header of the contribution
\begin{abstract}
The pandemic of these very recent years has led to a dramatic increase in people wearing protective masks in public venues. This poses obvious challenges to the pervasive use of face recognition technology that now is suffering a decline in performance. One way to address the problem is to revert to face recovery methods as a preprocessing step. Current approaches to face reconstruction and manipulation leverage the ability to model the face manifold, but tend to be generic. We introduce a method that is specific for the recovery of the face image from an image of the same individual wearing a mask. We do so by designing a specialized GAN inversion method, based on an appropriate set of losses for learning an unmasking encoder. With extensive experiments, we show that the approach is effective at unmasking face images. In addition, we also show that the identity information is preserved sufficiently well to improve face verification performance based on several face recognition benchmark datasets.

\end{abstract}

\input{main_intro}
\input{main_relevant_works}
\input{main_method}
\input{main_experiments}

\section{Conclusions}

In this work we have proposed a method for unmasking the face image of a subject wearing a mask. We formulate the problem as a GAN inversion, because we leverage the generative modeling of the face manifold of current methods. We designed a set of losses to learn an unmasking encoder that enables mapping the input image onto a new face image. Our set of experiments show that the unmasking process recovers compelling face images, with competitive image quality metrics. In addition, by testing the unmasking process with two face matchers, our set of results on face verification confirms that the identity is preserved sufficiently well to provide a consistent significant improvement on three commonly used face recognition benchmarks.

\subsubsection{Acknowledgements} This material is based upon work supported in part by the Center for Identification Technology Research and the National Science Foundation under Grants No. 1650474 and No. 1920920.

\bibliographystyle{splncs04}
\bibliography{main_references}

\end{document}

%% file: main_intro.tex
\section{Introduction}

Face recognition in unconstrained environments is still a challenging problem, despite the impressive progress of recent approaches based on deep learning~\cite{Schroff2015-bt,Deng2018-xp}. A major factor affecting performance is the presence of occluded parts of the face. Although face recognition under occlusions is not a new problem~\cite{Zeng_19}, its relevance has been refreshed in light of the COVID-19 pandemic, which has led to a dramatic increase of people wearing protective masks of various kinds in public venues. This new \emph{status quo} is posing challenges to the pervasive use of face recognition technology, leading to government institutions initiating studies to better evaluate the effects of face masks on current approaches~\cite{Ngan2020-bo}.

There is more than one way to mitigate the loss of performance of face matchers dealing with face images wearing masks~\cite{Zeng_19}, and one of them is to attempt to reconstruct the face appearance on the occluded region. The main advantage of this approach is that it can be used to potentially improve the performance of any face matcher.

Recent approaches for face reconstruction and manipulation based on deep learning~\cite{Donahue2016-lp,Richardson_72,Pidhorskyi_70} leverage the extraordinary generative power of these methods in capturing the statistics of the face manifold~\cite{karras_71}. In this work we plan to harness that capability even further. Differently than previous approaches, which aim at generic face manipulations, we develop a method that is specifically focussed on unmasking images of faces wearing masks. Our method does not involve the detection or segmentation of face masks, and can be used as a preprocessing step to unmask a face image, which can then be fed to a face matcher.

We frame the problem as a special instance of a GAN inversion~\cite{Daras_47,Roich_48}, where the GAN network is a StyleGAN2 architecture~\cite{karras_71}. We do so by designing a set of losses and a training procedure for learning an encoder network that maps the input image of a face wearing a mask onto an appropriate code space of faces not wearing masks. This is meant to be the input space of the generator network that will then reproduce the face image without mask.

Ultimately, the challenge is to generate face images that preserve the identity of the input in order to improve face recognition performance. This is why we test our approach with several face recognition datasets. In particular, we show that it can produce compelling face reconstructions with competitive image quality metrics. In addition, we evaluate extensively how our method works for improving face verification under several face masking conditions.

%%% Local Variables:
%%% mode: latex
%%% TeX-master: "main"
%%% End:

%% file: main_relevant_works.tex
\section{Relevant Works}

\textbf{Image recovery under occlusion and recognition}. Our work can be considered as a type of occlusion recovery which coud be used for face recognition~\cite{Zeng_19}. A lot of works treat occlusions as noise and compress the occluded images/faces down to a lower resolution or latent space. This helps to filter out the noise and reconstruct the images back at a higher resolution. Some of these approaches employ more traditional methods like sparse representations~\cite{Su_25,Li_17,Fidler_30,Luan_32,Iliadis_33,Leonardis_34,Zhao_38,Iliadis_41} and PCA~\cite{Park_10,Deng_29,Luan_32}, while others rely on neural networks~\cite{Li_5,Gao_14,He_31,Xie_102} to accomplish this task. Our approach is more similar to the latter. Additionally, some of these methods are occlusion aware and rely on occlusion segmentation or contours to help with the image recovery process~\cite{Din_2,Xiong_3,Wang_9,Chen_16,Xinyi_36,Abdal_57}. Therefore, occlusion map prediction is part of their model. On the other hand, we do not require any kind of occlusion information. Additionally, some works try to make occlusion neutral feature-extractors/encoders \cite{Liu_7,Cheng_22} or train the face matching networks to adapt to occlusions~\cite{Trigueros_101}. In that sense, our method is not occlusion neutral, and although we do employ ArcFace~\cite{Deng2018-xp} and FaceNet~\cite{Schroff2015-bt} to help with the training and facial verification tasks, at no point do we train these matchers.

\textbf{GAN inversion}. A lot of approaches used local discriminators and global discriminators and trained GANs from scratch to reproduce faces/images free of occlusions~\cite{Iizuka_6,Ge_26,Li_39}. Some others used variations of cyclic losses for image/facial deocclusion~\cite{Hu_4,Li_20}. We decided to use a pretrained StyleGAN2~\cite{karras_71} as our generator. As stated in~\cite{Abdal_59,Zhu_60}, real life face reconstruction via StyleGAN based on the original $\mathcal{W}$ space is a very hard task. Some approaches have extended the $\mathcal{W}$ space to new ones, named $\mathcal{W}+$~\cite{Wei_65,Richardson_72}, $\mathcal{W}*$~\cite{Shukor_55}, $p$~\cite{Zhu_60} etc, while others have trained the decoders with various losses to achieve exact facial GAN inversion~\cite{Daras_47,Roich_48,Ghosh_54}. We decided to build our approach based on the $\mathcal{W}+$ space via the pSp model~\cite{Richardson_72} and keep the generator fixed. Additionally, related to us, a couple of works employ $\ell$-norm losses in the latent space of their StyleGAN architectures~\cite{Pidhorskyi_70,Ghosh_54,Nitzan_56}. However, none of them use the latent space loss for image recovery or inpainting with StyleGAN.

We also report that~\cite{Li_39,Pernus_53,Wei_65} employed face parser losses in the output space while training for GAN inversion/facial reconstruction). However, our work does not use such a loss. Perceptual losses like LPIPS were also used by us and other works~\cite{Xinyi_36,ma2022contrastive,Wei_65}. ID losses based on Facenet, ArcFace, LightCNN etc were also used while training for GAN inversion/face unmasking~\cite{Ge_26,Xinyi_36,Zhao_37,Nitzan_56,Wei_65,ma2022contrastive}. Just like~\cite{Li_20,Li_39,Menon_46,Wei_65,ma2022contrastive}, we also employed an $\ell$-norm loss on the output image space to help with image reconstruction.

%%% Local Variables:
%%% mode: latex
%%% TeX-master: "main"
%%% End:

%% file: main_method.tex
\section{Method}

Given an image $M$ of a face wearing a mask, we are interested in developing an approach for face unmasking, which is the task of mapping $M$ onto a new image $U$, depicting the same person in $M$, only without the mask. We assume that the unmasking process can be modeled by the relationship $U = g \circ f (M)$, where $f$ maps $M$ onto a representation $w$, and $g$ generates $U$ from the representation. We do not make assumptions about the specific type of face mask, nor do we require a mask detection or segmentation process to be involved in the unmasking task. We do however, require the face in $M$ to be aligned in terms of 2D position, 2D orientation and scale with the nominal alignment of the dataset used for training the model $g \circ f$.   

\subsection{Baseline Model}

In the case when the face in the image $M$ was not wearing a mask, since no mask needs to be removed, we would expect this condition to be true: $U = M$. Also, let us indicate with $T \doteq U = M$ the image of the face without mask.  Therefore, the model $g \circ f_0$ should behave like a face autoencoder, where the encoder in this particular case is indicated with $f_0$. While there are several implementations of face autoencoders~\cite{Donahue2016-lp,Richardson_72,Pidhorskyi_70}, since we are ultimately interested in evaluating how the approach would improve the performance of face recognition, we want one that executes face autoencodings that are photorealistic, and that can maintain face identity. The state-of-the art in that category is the pSp model~\cite{Richardson_72}, where the generator $g$ is a StyleGAN2 network~\cite{karras_71}, and the encoder $f$ is based on a feature pyramid model built on top of a ResNet backbone, and followed by a mapping to a set of 18 \emph{styles}. The styles capture different levels of image detail, roughly divided in three groups, coarse, medium, and fine. Every style is a 512-dimensional vector. The collection of the 18 style vectors constitutes the representation $w$, which is an element of the space referred to as $\mathcal{W}+$ in~\cite{Richardson_72}.

The training of the pSp model is approached as a ``GAN inversion'' task, meaning that the generator network is trained offline (i.e., StyleGAN2), and is kept locked while only the encoder $f_0$ is being trained, with the task of ``learning  to invert'' the operation of the generator. This approach is mainly due to the success of StyleGAN2 in modelling the face space, and also due to the difficulty in designing and training such kind of models.

In order to train the encoder $f_0$, the pSp model combines a number of losses. The fist one is a reconstruction loss based on the $\ell_2$-norm
\begin{equation}
  \mathcal{L}_R (T) = \| T - g\circ f_0(T) \|_2 \; .
\end{equation}
The second aims at maintaining the perceptual similarity between input and reconstructions, and is based on the LPIPS metric $P(\cdot)$~\cite{Zhang2018-oq}
\begin{equation}
  \mathcal{L}_{LPIPS} (T) = \| P(T) - P(g\circ f_0(T)) \|_2 \; .
\end{equation}
In order to preserve the face identity of the input in the reconstructions, an identity preserving loss is used to maximize the cosine similarity between the normalized  ArcFace~\cite{Deng2018-xp} representations $AF(\cdot)$ of the input image and the reconstruction
\begin{equation}
  \mathcal{L}_{ID} (T) = 1-AF(T)\cdot AF(g\circ f_0(T)) \; .
\end{equation}

The encoder $f_0$, which we refer to as the \emph{baseline encoder}, is then trained by minimizing this loss, which is written on a per-image basis as
\begin{equation}
  \mathcal{L}_0 (T) =  \mathcal{L}_R (T) + \alpha \mathcal{L}_{LPIPS} (T) + \beta \mathcal{L}_{ID} (T) \; , % + \gamma \mathcal{L}_w (T) \; ,
  \label{eq-loss-psp}
\end{equation}
where $\alpha$, and $\beta$ are hyperparameters striking a balance between the loss terms.
\begin{figure}[t!]
	\centering
	\includegraphics[width=\textwidth]{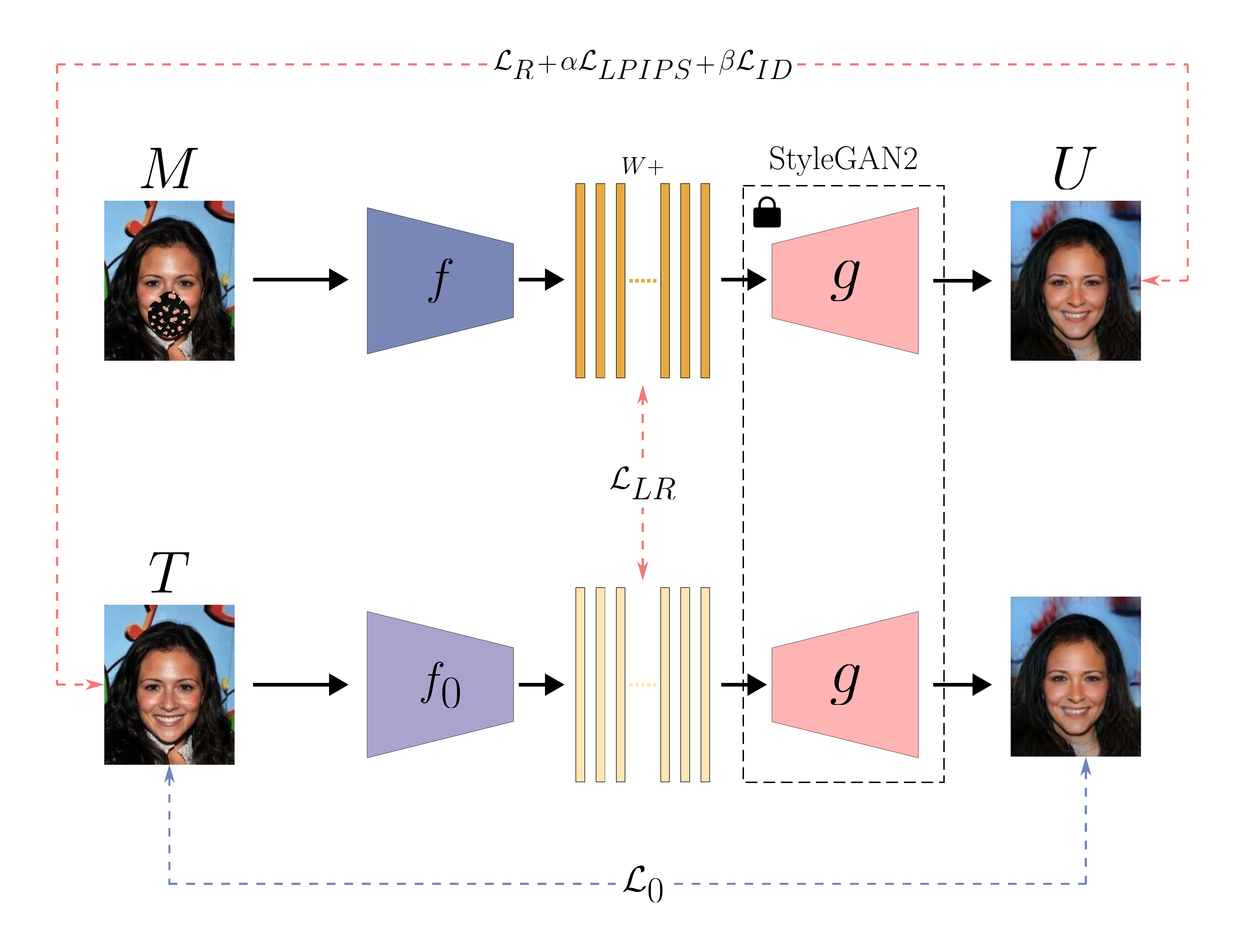}
	\caption{\textbf{Unmasking Architecture.} Overview of the architecture and the losses used to train the baseline encoder $f_0$, and the unmasking encoder $f$. The generator $g$ is kept fixed at all times. $T$ is a face image. $M$ is the same face image wearing a mask. $U$ is the autoencoded version of $M$.}
	\label{fig-example}
\end{figure}
      
\subsection{Unmasking Model}

Given an image $M$ with a masked face, we can still make the assumption that in our original model $g \circ f$, $g$ is a face image generator, modeled with StyleGAN2, and that we keep it fixed. Therefore, training the encoder $f$ becomes a specialized GAN inversion problem. If $T$ is an image of a face not wearing a mask, we make the assumption that $T$ is identical to $M$, except for the area of $M$ corresponding to the pixels on the face mask.

To train the encoder $f$ we combine several losses, most of which are a modification of those used to train the baseline model $f_0$. See Figure~\ref{fig-example}. Specifically, we require the autoencoding of $M$ (i.e., $U$) to be close to $T$ in the $\ell_2$-norm sense
\begin{equation}
  \mathcal{L}_R(T,M) = \| T - g\circ f(M) \|_2 \; .
  \label{eq-loss-r}
\end{equation}
We also want the autoencoding of $M$ to be perceptually similar to $T$ according to the LPIPS metric $P(\cdot)$ by minimizing
\begin{equation}
  \mathcal{L}_{LPIPS} (T,M) = \| P(T) - P(g\circ f(M)) \|_2 \; .
  \label{eq-loss-lpips}
\end{equation}
In addition, the identity of the autoencoding of $M$ should be as close to the identity of $T$ as possible, and for that we maximise the similarity between the respective normalized ArcFace representations $AF(\cdot)$ by minimizing
\begin{equation}
  \mathcal{L}_{ID}(T,M) = 1-AF(T)\cdot AF(g\circ f(M)) \; .
  \label{eq-loss-id}
\end{equation}

We also observe that ideally, the baseline model should be such that $T = g \circ f_0(T)$. Therefore, we would want as much as possible that $g \circ f_0(T) = g \circ f(M)$, but this could be achieved by simply having $f_0(T) = f(M)$. So, we encourage that with the loss
\begin{equation}
  \mathcal{L}_{LR} (T,M) = \| f_0(T) - f(M) \|_2 \; ,
  \label{eq-loss-lr}
\end{equation}
which we name \emph{latent reconstruction loss}. Finally, the encoder $f$, which we refer to as the \emph{unmasking encoder}, is trained by minimizing this combined \emph{unmasking loss}, which is written on a per-image basis as
\begin{equation}
  \mathcal{L}_{UM} (T,M) =  \mathcal{L}_R (T,M) + \alpha \mathcal{L}_{LPIPS} (T,M) + \beta \mathcal{L}_{ID}(T,M) + \gamma \mathcal{L}_{LR} (T,M) \; ,
  \label{eq-loss-um}
\end{equation}
where $\alpha$, $\beta$, and $\gamma$ are hyperparameters striking a balance between the loss terms.

\subsection{Datasets}

Because of the pandemic, there is a number of datasets and tools to add masks to face images. For instance, Masked-FaceNet is a dataset with faces from the FFHQ dataset wearing masks correctly and incorrectly~\cite{Cabani_66}, RMFD is a collection of real world masked faces, including also face images with same identity not wearing masks~\cite{Wang_68}, and so is DS-IMF~\cite{Mishra_69}. The MAFA dataset~\cite{Ge_67} has a lot of real world masked images, but neither with identification information nor with corresponding identities without wearing masks. In our experiments we used the FFHQ dataset~\cite{Karras2019-ra}, CelebA~\cite{liu2015faceattributes}, and LFW~\cite{Huang2012a}, and we used the MasktheFace toolkit~\cite{Anwar_45} to create the pairs of face images $(T,M)$, where $M$ is a version of $T$ with a synthetic mask added. Note that the MasktheFace toolkit failed to mask faces of certain images and those were not included in further training and testing of the models. In addition, we used a subsection of RFRD called RMFRD, which has real life masks only. Table~\ref{table:1} gives details on the size of the datasets used. 
\begin{table}[t!]
	\centering
	\caption{\textbf{Datasets.} Quantitative summary of the datasets used.}
		\begin{tabular}{|c| c |c |c| c|} 
			\hline
			Name & \thead{Original \# \\ of images} & \thead{Total \# \\ of masked images} & \thead{\# of train, \\ test IDs} & \thead{\# of train,\\ test images}\\ [0.5ex] 
			\hline
			FFHQ & 70000 & 69794 & ---,--- & 55811, 13593\\ 
			CelebA & 202599 & 196999 & 8141, 2036 & 157597, 39402 \\
			LFW &  13233 & 13168 & 4754, 1144 & 10794, 2374\\
			RMFRD & 2118 M + 90468 & 806 & 218, 64 & 597, 209 \\ [1ex] 
			\hline
		\end{tabular}
	\label{table:1}
      \end{table}

\subsection{Implementation Details}

For training our approach, we assume that a StyleGAN2 generator model $g$ is given to us and is kept locked. Then, we train the baseline encoder $f_0$ with the loss~\eqref{eq-loss-psp}. Subsequently, we use $f_0$ to initialize the unmasking encoder $f$, and we train it with the loss~\eqref{eq-loss-um}. Also, we conduct the experiments by first learning the model for the dataset with higher resolution, and then we use the baseline model to initialize the baseline model of the dataset with the immediate smaller resolution. So, we start from the FFHQ dataset~\cite{Karras2019-ra}, then we process CelebA~\cite{liu2015faceattributes}, then LFW~\cite{Huang2012a}, and finally RMFRD~\cite{Wang_68}.

The generator $g$ based on StyleGAN2 allows to generate images at $1024 \times 1024$ resolution with an architecture based on 18 layers. It is also possible to use only the first 14 layers of StyleGAN2 and work with a model that generates images at  $256 \times 256$ resolution~\cite{Richardson_72}, which is also much faster to train. We verified, as reported in Section~\ref{sec-experiments}, that working with the smaller network does not affect face verification results significantly, since the images are downscaled before feeding them to the face matcher. Therefore, unless otherwise stated, we always use the model with the smaller 14 layers generator architecture.

The approach we use assumes that the input images have faces that are sufficiently aligned. This is the case for FFHQ and CelebA. For LFW, we used the deep funneled images~\cite{Huang2012a}, which correct for the orientation of the faces and properly align them. We also crop a $150 \times 150$ region out of the original $250 \times 250$ images to leave out image areas containing significant background clutter that were making the training difficult to converge. Note also that the cropped images were then resized up to $256 \times 256$ prior to be used.

Additionally, RMFRD has very low quality images and a very variable resolution, and it is in general a very challenging dataset. Because of this, we used OpenCV to only keep faces where we could detect both eyes, and we rotated the faces to make the eyes horizontal, and resize them to $256 \times 256$. Moreover, this dataset has faces wearing real masks as $M$ images, and there are no identical images with faces without masks as $T$ images. Therefore, the training is approached in two phases. First, we used the unmasked faces in the training set to train the baseline and the unmasking models just like we did for the other datasets. Second, we fine-tune the model with the images with real masks as follows. The losses~\eqref{eq-loss-r} and~\eqref{eq-loss-lpips} are computed based only on the periorbital region of the face because it is visible, which is identified automatically from the position of the eyes. The loss~\eqref{eq-loss-id}, instead, uses as $T$, an image with the same identity and that is not wearing a mask. Finally, in the loss~\eqref{eq-loss-lr} $T$ is replaced with an estimate $\hat T$ of a face image without mask, generated by $g$ to have the same periorbital region of the masked face in the $M$ image according to the $\ell_2$-norm.

%%% Local Variables:
%%% mode: latex
%%% TeX-master: "main"
%%% End:

%% file: main_experiments.tex
\begin{figure}[!t]
	\centering
	\begin{minipage}{.49\textwidth}
		\centering
		\begin{subfigure}[h]{1.0\textwidth}
			\centering
			\includegraphics[width=\textwidth]{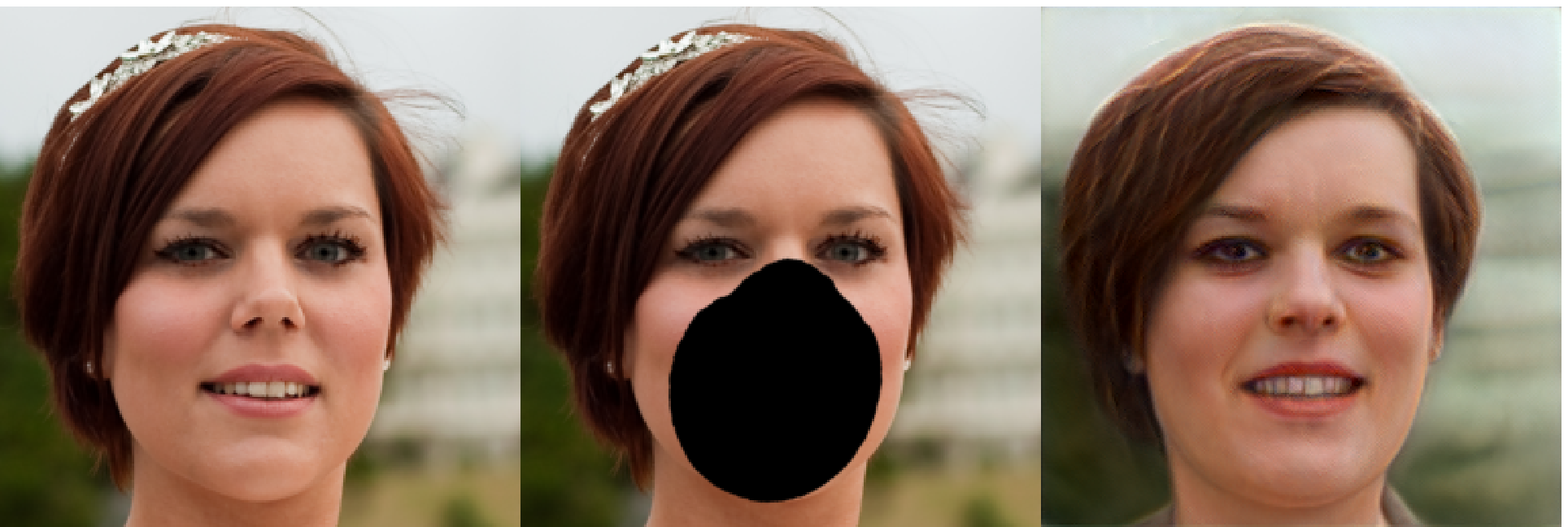}
		\end{subfigure}
		\begin{subfigure}[h]{1.0\textwidth}
			\centering
			\includegraphics[width=\textwidth]{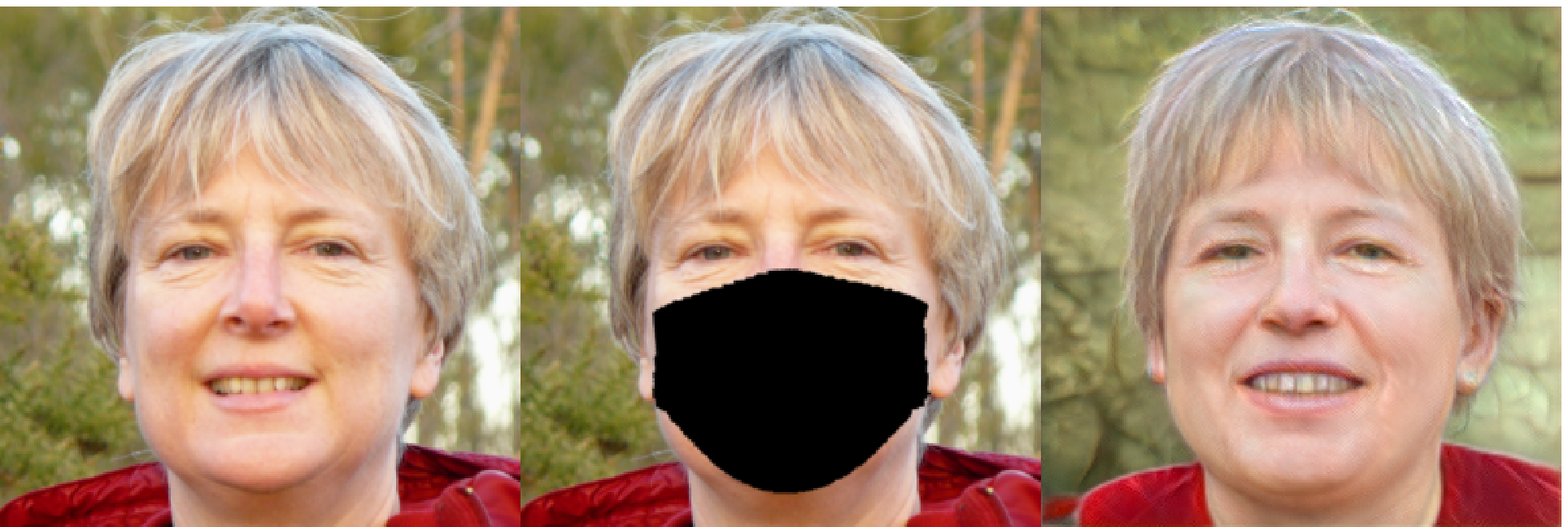}
		\end{subfigure} \\
		(a) FFHQ
	\end{minipage}
	\begin{minipage}{.49\textwidth}
		\centering
		\begin{subfigure}[h]{1.0\textwidth}
			\centering
			\includegraphics[width=\textwidth]{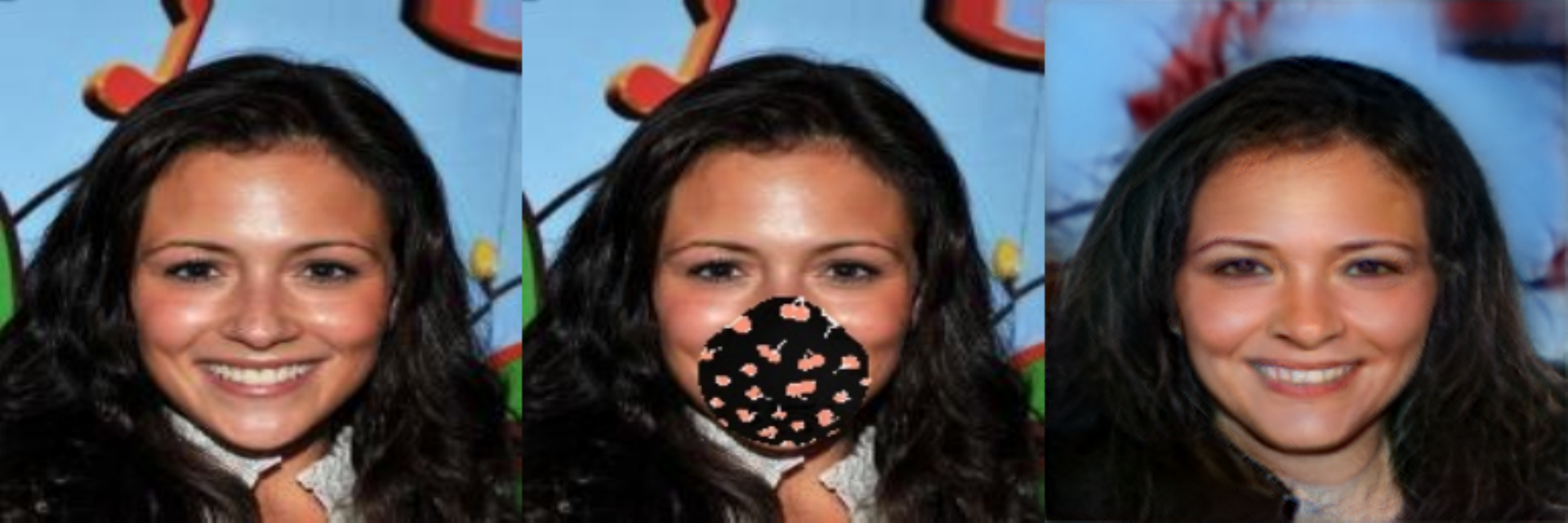}
		\end{subfigure}
		\begin{subfigure}[h]{1.0\textwidth}
			\centering
			\includegraphics[width=\textwidth]{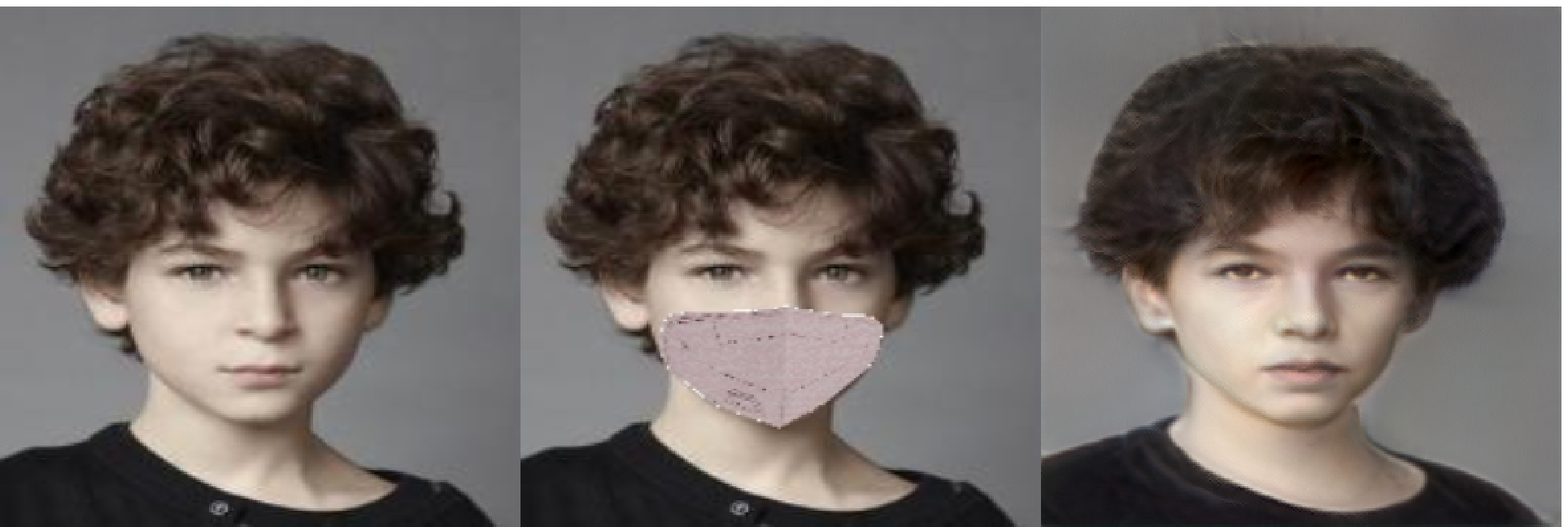}
		\end{subfigure}
		(b) CelebA
	\end{minipage}
	
	\begin{minipage}{.49\textwidth}
		\centering
		\begin{subfigure}[h]{1.0\textwidth}
			\centering
			\includegraphics[width=\textwidth]{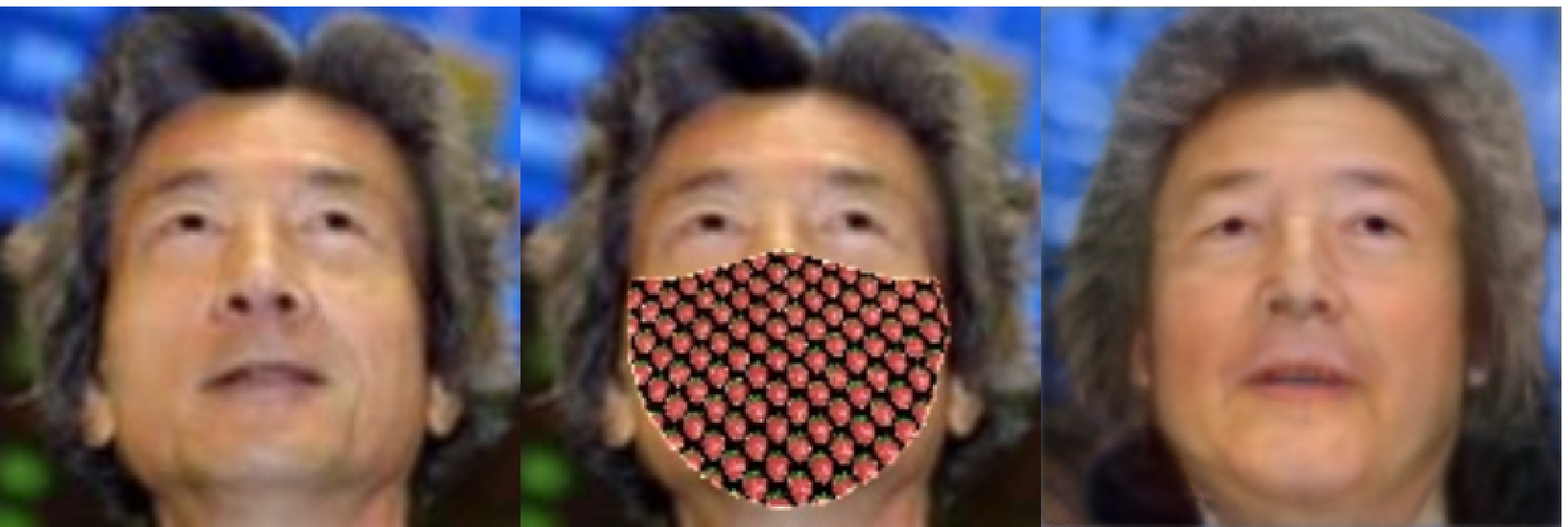}
		\end{subfigure}
		\begin{subfigure}[h]{1.0\textwidth}
			\centering
			\includegraphics[width=\textwidth]{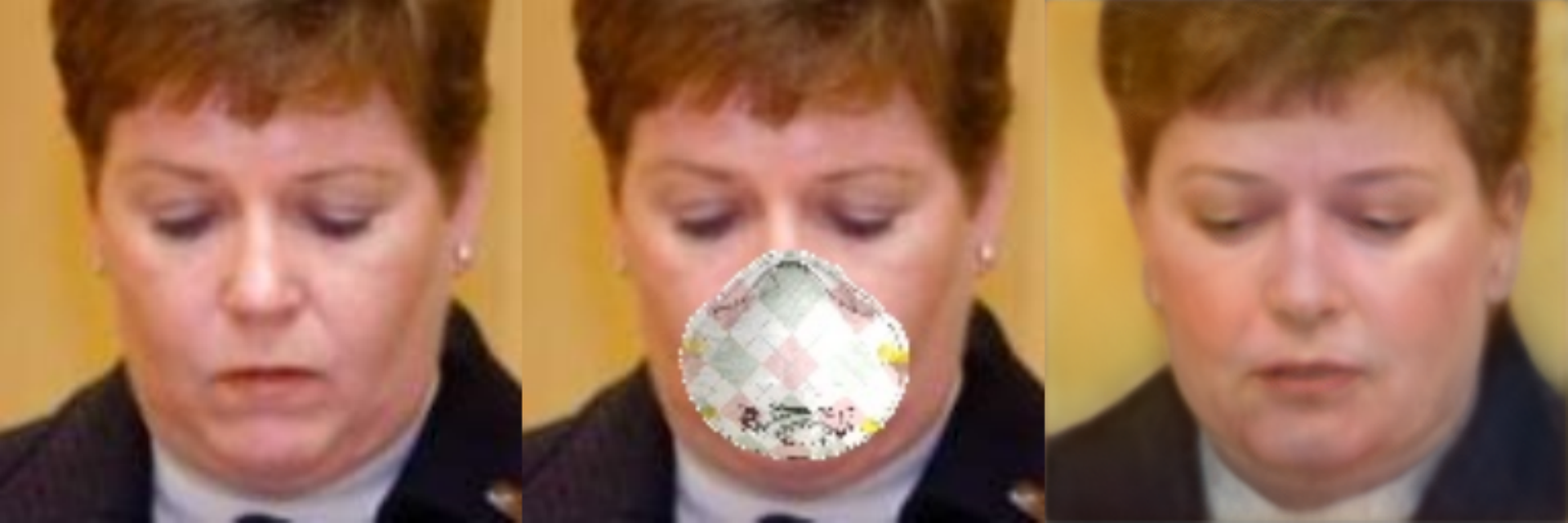}
		\end{subfigure}
		(c) LFW
	\end{minipage}
	\begin{minipage}{.49\textwidth}
		\centering
		\begin{subfigure}[h]{1.0\textwidth}
			\centering
			\includegraphics[width=\textwidth]{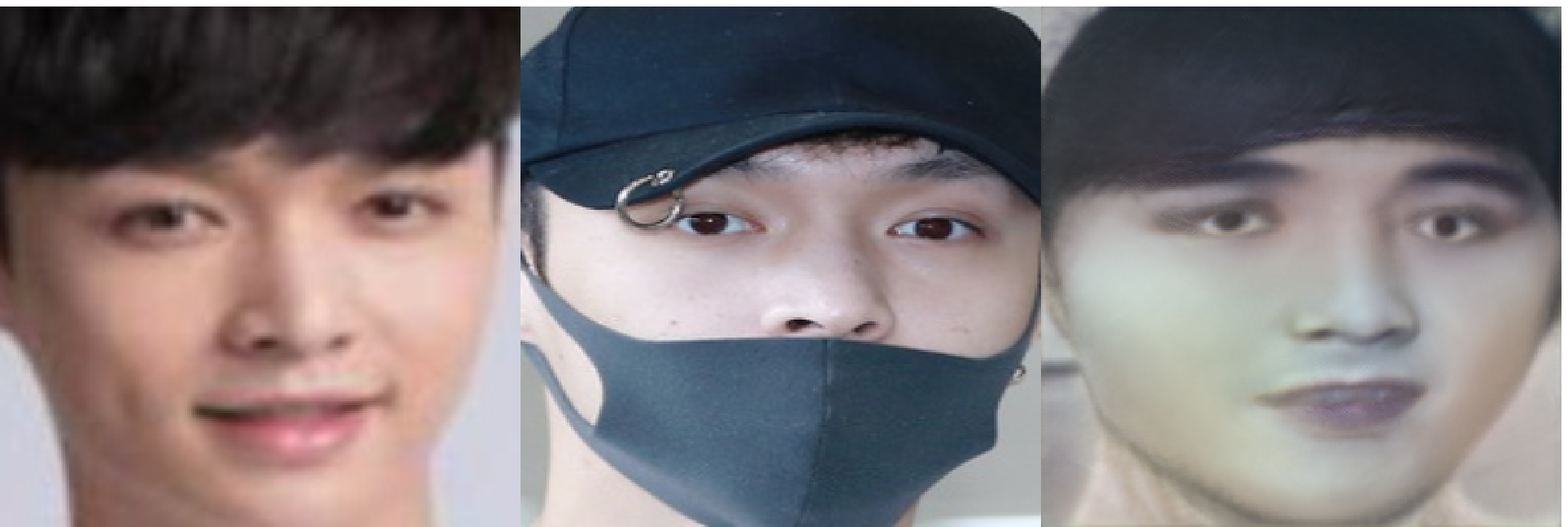}
		\end{subfigure}
		\begin{subfigure}[h]{1.0\textwidth}
			\centering
			\includegraphics[width=\textwidth]{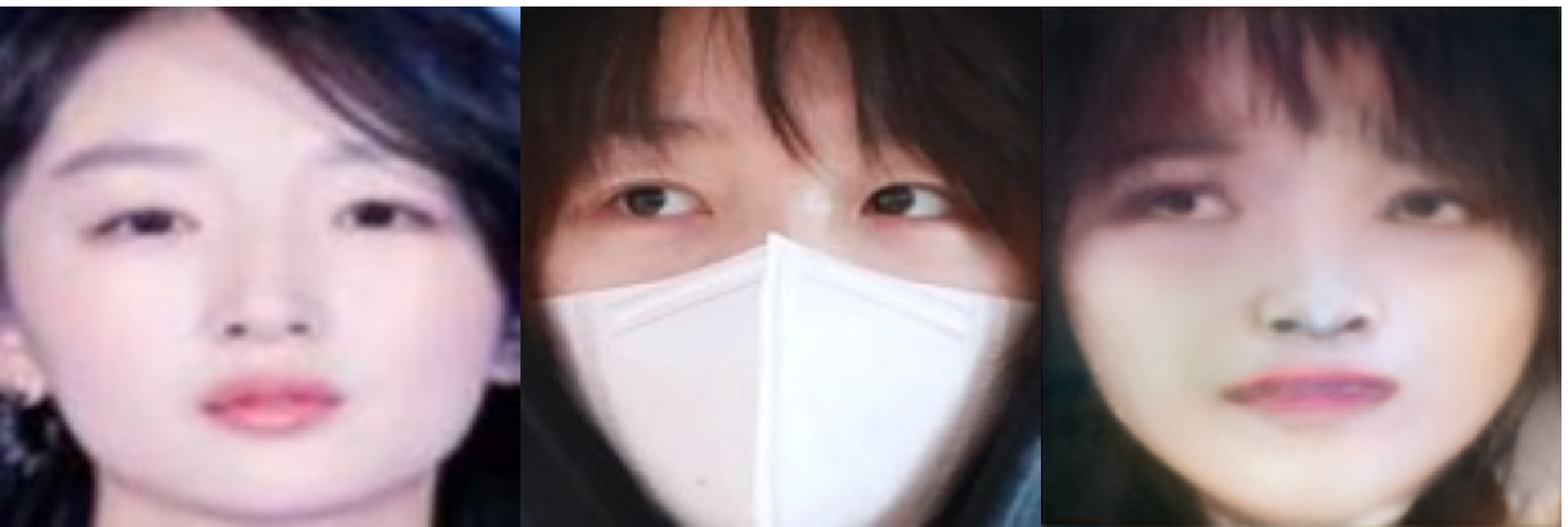}
		\end{subfigure}
		(d) RMFRD
	\end{minipage}
	\caption{\textbf{Face unmasking.} Qualitative face unmasking results on various datasets. For each row we have original face image ($T$), masked image ($M$), and unmasked image ($U$).}
	\label{fig:unmask-results}
\end{figure}

\section{Experimental Results}
\label{sec-experiments}

We evaluate our unmasking model extensively, by providing results pertaining the unmasking of face images wearing masks, the image quality metrics of those images, and we evaluate to what extent the unmasking process might help improving the performance of a face matcher. We use four datasets, FFHQ~\cite{Karras2019-ra}, CelebA~\cite{liu2015faceattributes}, and LFW~\cite{Huang2012a}, and RMFRD~\cite{Wang_68}, and two face matchers, ArcFace\cite{Deng2018-xp} and FaceNet~\cite{Schroff2015-bt}.

\textbf{Face verification notation.} The face verification experiments are conducted with different unmasking settings. The notation used to indicate these settings is defined as follows. MM indicates when both the probe and the gallery face images are masked. MT indicates when the probe image is masked and gallery images are not. UU indicates that both probe and gallery images were originally masked but they were both unmasked by our approach before verification. UT indicates that the probe was originally masked but was unmasked by our approach before verification and the gallery images were not masked. TT indicates when the probe and the gallery images were not masked. This is expected to provide the upper bound results.

\textbf{18 layers vs 14 layers architecture.} We compared the two generator architectures based on 14 and 18 layers in terms of face verification performance as well as image quality of the unmasked images. A key difference between the architectures is that the 14 layers provides $256 \times 256$ images and the 18 layers provides $1024 \times 1024$ images.  Table~\ref{table:pSpDepths} shows the results on CelebA. The metrics used are area under curve (AUC) for face verification, and peak signal to noise ratio (PSNR) and structural similarity index measure (SSIM) for image quality~\cite{Hore2010-tb}. The main conclusion is that the difference between architectures is not significant. For AUC purposes, this is not surprising because the images are downsampled to $112 \times 112$ for the face matchers. Also, the pSp framework only accepts input at a resolution of $256 \times 256$. Therefore when calculating SSIM and PSNR, the $1024 \times 1024$ images are downsized to $256 \times 256$ for the model input and the resulting output quality is very similar to that of the 14 layers model. The 18 layer model has slightly better facial verification AUCs because it has more expressive power. On the other hand, the 14 layer model has marginally better SSIM and PSNR values because it has to produce images at a lower scale. Therefore, in the remaining experiments we used the 14 layered architecture, which also allows for a faster training time.

\begin{table}[t!]
	\centering
	\caption{\textbf{Architecture Depth.} Effects of the generator network depth on face verification in terms of AUC, and unmasking image quality in terms of PSNR and SSIM. The dataset used is CelebA. }
	\begin{tabular}{c c c c c} 
		Architecture & UU & UT & PSNR & SSIM \\ [0.5ex] 
		\hline
		14-layer  & 0.9590 & 0.9422 & 19.00 & 0.75891 \\
		18-layer  & 0.9596 & 0.9440 & 18.51 & 0.75568 \\
	\end{tabular}
	\label{table:pSpDepths}
\end{table}

\textbf{Face unmasking.} We conducted face unmasking experiments with the four datasets FFHQ, CelebA, LFW, and RMFRD. Qualitative unmasking results can be seen in Figure~\ref{fig:unmask-results}. The unmasked images have resolution $256 \times 256$. For FFHQ the mask was simply black, for CelebA and LFW we show the case of different shape and color masks added, and the masks in RMFRD are real masks worn by the subjects.
\begin{table}[t!]
	\centering
 	\caption{\textbf{Face verification.} AUC, SSIM, and PSNR values of an ablation study of the unmasking loss for different face verification settings. ArcFace was used as face matcher.}
	\begin{tabular}{| c | c | c | c | c | c | c | c | c | c |}
		\hline
		Dataset & $\mathcal{L}_R$ & $\mathcal{L}_{LPIPS}$ & $\mathcal{L}_{ID}$ & $\mathcal{L}_{LR}$ & \diagbox{MM}{UU} & \diagbox{MT}{UT} & TT & SSIM & PSNR\\ [0.5ex] 
		\hline\hline
		FFHQ & $\checkmark$ & $\checkmark$ & $\checkmark$ & $\checkmark$ & & &  & 0.69450 & 17.92 \\
		\hline\hline
		 CelebA & $\checkmark$ &  &  &  & \diagbox{0.825}{0.750} & \diagbox{0.824}{0.750} & 0.891 &  & \\
		 \hline
         CelebA &  &  &  & $\checkmark$ & \diagbox{0.825}{0.833} & \diagbox{0.824}{0.840} & 0.891 & 0.72905 & 18.30\\
         \hline
		CelebA & $\checkmark$ & $\checkmark$ & $\checkmark$ &  & \diagbox{0.931}{0.952} & \diagbox{0.940}{0.964} & 0.984 & & \\
		\hline
		CelebA &  & $\checkmark$ & $\checkmark$ & $\checkmark$ & \diagbox{0.931}{0.947} & \diagbox{0.940}{0.962} & 0.984 & & \\
		 \hline
		CelebA & $\checkmark$ & $\checkmark$ & $\checkmark$ & $\checkmark$ & \diagbox{0.931}{0.959} & \diagbox{0.940}{0.971} & 0.984 & 0.75891 & 19.00\\
		\hline\hline
		LFW & $\checkmark$ & $\checkmark$ & $\checkmark$ &  & \diagbox{0.952}{0.944} & \diagbox{0.958}{0.958} & 0.990 & & \\
		\hline
		LFW & $\checkmark$ & $\checkmark$ & $\checkmark$ & $\checkmark$ & \diagbox{0.952}{0.957} & \diagbox{0.958}{0.968} & 0.990 & 0.67737 & 17.04\\ 
		\hline\hline
		RMFRD & $\checkmark$ & $\checkmark$ & $\checkmark$ & $\checkmark$ & & \diagbox{0.602}{0.609} & & & \\
		\hline
	\end{tabular}
	\label{table:AUCs}
      \end{table}

      \textbf{Ablation of the unmasking loss.} In Table~\ref{table:AUCs} we report an ablation study where in the unmasking loss~\eqref{eq-loss-um} we include only the components indicated. We do so for a face verification experiment using the CelebA, LFW, and RMFRD datasets. The AUC values highlight the contribution coming from  using only $\mathcal{L}_{R}$, only $\mathcal{L}_{LR}$, and how much performance deteriorates when each of them is removed from the full model. 
      
      The first two experiments in Table~\ref{table:AUCs}
      concentrate on the cases when the model is only trained with either $\mathcal{L}_{R}$ or $\mathcal{L}_{LR}$. The facial verification results for these two models only use 3000 images from the test dataset (the rest of the models use the entirety of the test dataset). Please note that $\mathcal{L}_{R}$ is not enough to learn a model where the UU case is better than the MM, or the UT case is better than the MT, highlighting the fact that adding the $\mathcal{L}_{LPIPS}$ and $\mathcal{L}_{ID}$ is important for $\mathcal{L}_{R}$. For CelebA, a model with only $\mathcal{L}_{LR}$ allows the UU and UT cases to outperform the MM and MT cases respectively, but adding the rest of the losses further increases the facial verification AUCs by a large margin. Additionally, for LFW, which is a lower resolution dataset, removing $\mathcal{L}_{LR}$ causes UU and UT to not outperform MM and MT respectively, further highlighting its importance. The final takeaway is that the full model allows for UU and UT to surpass MM and MT, respectively, and allows to approach the upper bound set by the TT scenario.
      
\begin{figure}[t!]
	\centering
	
	\begin{minipage}{1.0\textwidth}
		\centering
		\begin{subfigure}[h]{0.16\textwidth}
			\centering
			\includegraphics[width=\textwidth]{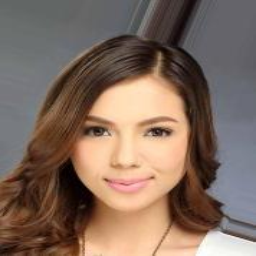} \\
			$T$
		\end{subfigure}
		\begin{subfigure}[h]{0.16\textwidth}
			\centering
			\includegraphics[width=\textwidth]{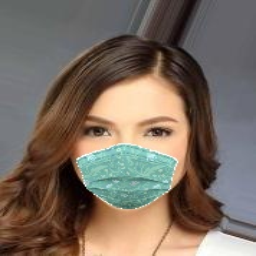} \\
			$M$
		\end{subfigure}
		\begin{subfigure}[h]{0.16\textwidth}
			\centering
			\includegraphics[width=\textwidth]{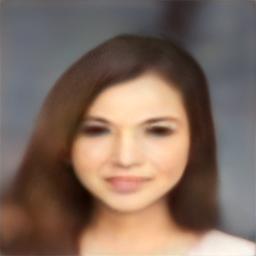} \\
			$U_{\eqref{eq-loss-r}}$
		\end{subfigure}
		\begin{subfigure}[h]{0.16\textwidth}
			\centering
			\includegraphics[width=\textwidth]{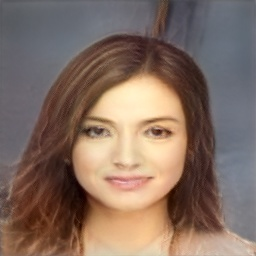} \\
			$U_{\eqref{eq-loss-lr}}$
		\end{subfigure}
		\begin{subfigure}[h]{0.16\textwidth}
			\centering
			\includegraphics[width=\textwidth]{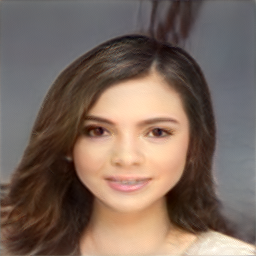} \\
			$U_{\eqref{eq-loss-r}+\eqref{eq-loss-lpips}+\eqref{eq-loss-id}}$
		\end{subfigure}
		\begin{subfigure}[h]{0.16\textwidth}
			\centering
			\includegraphics[width=\textwidth]{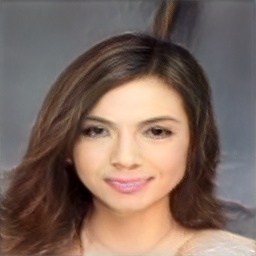} \\
			$U_{\eqref{eq-loss-um}}$
		\end{subfigure}

	\end{minipage}

	\caption{\textbf{Ablation of face unmasking.} Left two images: original face image ($T$) and same image with mask ($M$). The right four images are unmasked versions of $M$ obtained with different models.}
	\label{fig:LessLosses}
\end{figure}

      Also, Table~\ref{table:AUCs} summarizes the results on image quality metrics such as SSIM and PSNR. LFW has lower metrics than the other datasets because it has the lowest quality images, especially due to the aforementioned cropping of the $T$, and $M$ images from a size of $250 \times 250$ to $150 \times 150$ followed by an upsampling to $256 \times 256$. Additionally, FFHQ also performs worse than CelebA because the comparison between $U$ and $T$ is done at a resolution of $256 \times 256$, for which the $T$ has to be downsampled, which detrimentally affects the SSIM and PSNR metrics. CelebA images, instead, are three times more than FFHQ for training, and are upsampled from $178 \times 218$ to $256 \times 256$, suffering the least amount of distortion. Finally, the CelebA model with only $\mathcal{L}_{LR}$ performs worse than the full model in terms of both SSIM and PSNR. This is because the latter has more constraints to satisfy, which gives better GAN-inversion results. More qualitative results can be seen in Figure~\ref{fig:LessLosses}, which follows similar trends.

Figure~\ref{fig:LessLosses}, instead, shows a qualitative ablation of the unmasking results, obtained by progressively adding more components in the unmasking loss~\eqref{eq-loss-um}. From the left, we have the original face image ($T$), and the version wearing the mask ($M$), followed by an unmasked face ($U$) with a model trained only with $\mathcal{L}_R$, which is rather blurry, despite the fact that the generator is a StyleGAN2 network that produces sharp face images. Then, fourth from the left, the unmasked face was obtained with a model trained only with $\mathcal{L}_{LR}$, which is relatively sharp, but the identity drift is noticeable. Second from the right, the unmasked face was obtained with a model without only $\mathcal{L}_{LR}$, whereas the last image was unmasked by the full model. 

\textbf{Face verification with FaceNet.} In Table~\ref{table:FaceNetAUCs} we report face verification results on CelebA and LFW based on FaceNet~\cite{Schroff2015-bt} as the face matcher. The train/test split used are the same as those used in the complete $\mathcal{L}_{UM}$ models in Table~\ref{table:AUCs}. Note that the results establish the same relationships between the various settings as those deductible from Table~\ref{table:AUCs}. This is relevant because now we have trained the unmasking models with one face matcher (i.e., ArcFace), while we have tested them with another one (i.e., FaceNet), confirming that even the previous results were not subject to strong biases, since models were trained and tested with the same face matcher, since ArcFace is used in the loss~\eqref{eq-loss-id}.
\begin{table}[h!]
	\centering
	\caption{\textbf{Face verification.} AUC face verification results on CelebA and LFW. FaceNet was used as face matcher during testing.}
	\begin{tabular}{| c | c | c | c | c | c | c |}
		\hline
		Dataset & $\mathcal{L}_R$ & $\mathcal{L}_{LPIPS}$ & $\mathcal{L}_{ID}$ & $\mathcal{L}_{LR}$ & \diagbox{MM}{UU} & \diagbox{MT}{UT} \\ [0.5ex]
		\hline\hline
		CelebA & $\checkmark$ & $\checkmark$ & $\checkmark$ & $\checkmark$ & \diagbox{0.832}{0.868} & \diagbox{0.814}{0.867} \\
		\hline\hline
		LFW & $\checkmark$ & $\checkmark$ & $\checkmark$ & $\checkmark$ & \diagbox{0.9834}{0.9833} & \diagbox{0.9844}{0.9845} \\
		\hline
	\end{tabular}
	\label{table:FaceNetAUCs}
 \end{table}

%%% Local Variables:
%%% mode: latex
%%% TeX-master: "main"
%%% End: